\documentclass[11pt,a4paper]{article}
\usepackage[hyperref]{acl2021}
\aclfinalcopy
\usepackage{times}
\usepackage{latexsym}

\usepackage{tabularx}

\usepackage{microtype}
\usepackage{graphicx}
\usepackage{xcolor}
\definecolor{darkgreen}{HTML}{197419}
\aclfinalcopy %
\def\aclpaperid{82} %

\title{Telling Stories through Multi-User Dialogue\\ by Modeling Character Relations}

\author{Wai Man Si \And Prithviraj Ammanabrolu  \\ School of Interactive Computing \\
  Georgia Institute of Technology \\
  \texttt{\{wssi, raj.ammanabrolu, riedl\}@gatech.edu} \\
  \And Mark O. Riedl \\
  }

\date{}

\begin{document}
\maketitle
\begin{abstract}
This paper explores {\em character-driven story continuation}, in which the story emerges through characters' first- and second-person narration as well as dialogue---requiring models to select language that is consistent with a character's persona and their relationships with other characters while following and advancing the story.
We hypothesize that a multi-task model that trains on character dialogue plus character relationship information improves transformer-based story continuation.
To this end, we extend the Critical Role Dungeons and Dragons Dataset~\citep{rameshkumar-bailey-2020-storytelling}---consisting of dialogue transcripts of people collaboratively telling a story while playing the role-playing game Dungeons and Dragons---with automatically extracted relationships between each pair of interacting characters as well as their personas.
A series of ablations lend evidence to our hypothesis, showing that our multi-task model using character relationships improves story continuation accuracy over strong baselines.
\end{abstract}

\section{Introduction}

Automated storytelling can be thought of as creative, long-from text generation and understanding---requiring explicit long-term memory, consistency, and creativity among other pre-requisites.
Most modern (neural) automated storytellers are {\em plot-driven} and frame the task in terms of sequentially generating plot points that narrate the story in third-person~\citep{kiros2015skip,Mostafazadeh2016c,martin2018,Fan2018}.
This approach does not generally place much weight on individual characters or their interactions---information known to be critical for creating stories~\citep{Riedl2010a}.

\begin{table}[h]
\footnotesize
\centering
\begin{tabular}{|rl|}
\hline
\multicolumn{1}{|l}{\textbf{Relations}} & 
\begin{minipage}{.7\linewidth}\begin{tabularx}{\linewidth}{X}
$\langle$ Scanlan, neutral, Vexahlia $\rangle$, \\ $\langle$ Keyleth, positive, Scanlan$\rangle$, \\ $\langle$ Grog, negative, Vexhalia$\rangle$, \\ $\langle$ Scanlan, positive, Vaxildan $\rangle$ ... \end{tabularx}\end{minipage}\\ \hline \hline
\multicolumn{1}{|l}{\textbf{Summary}}        & 
\begin{minipage}{0.7\linewidth}\begin{tabularx}{\linewidth}{X}
They wake up in the morning, preparing for the coming battle.  Scanlan turns them all into Ravenites with light clothing. The sleet storm is starting. ...\end{tabularx}\end{minipage}                \\ \hline \hline
\textbf{Vexahlia:}                                    & \begin{minipage}{0.7\linewidth}\begin{tabularx}{\linewidth}{X}  Bundle up! \end{tabularx}\end{minipage}                                                                                                                                                                   \\
\textbf{Scanlan:}                                    & 
\begin{minipage}{0.7\linewidth}\begin{tabularx}{\linewidth}{X}
 Okay. How will we know when it's time for me to release? We have to wait for Tooma to go report.\end{tabularx}\end{minipage}              \\
\textbf{Vexahlia:}                                    &
\begin{minipage}{0.7\linewidth}\begin{tabularx}{\linewidth}{X}
 Is Vorugal back? He's back.\end{tabularx}\end{minipage}                                                                                                                                              \\
\textbf{Scanlan:}                                   & 
\begin{minipage}{0.7\linewidth}\begin{tabularx}{\linewidth}{X}
 I assume.\end{tabularx}\end{minipage}                                                                                                                                                                   \\
\textbf{Vexahlia:}                                    & 
\begin{minipage}{0.7\linewidth}\begin{tabularx}{\linewidth}{X}
 \textcolor{blue}{\underline{Do we see Larkin around?}} \end{tabularx}\end{minipage}                                                                                                                                                  \\
\textbf{DM:}                                    & 
\begin{minipage}{0.7\linewidth}\begin{tabularx}{\linewidth}{X}
 \textcolor{blue}{\underline{No, you do not see Larkin around.}}\end{tabularx}\end{minipage}                                                                                                                                    \\
\textbf{Scanlan:}                                   & 
\begin{minipage}{0.7\linewidth}\begin{tabularx}{\linewidth}{X}
 Vax , do you want to go look?\end{tabularx}\end{minipage}                                                                                                                                               \\
\textbf{Vaxildan:}                                    & 
\begin{minipage}{0.7\linewidth}\begin{tabularx}{\linewidth}{X}
 For Larkin? \end{tabularx}\end{minipage}                                                                                                                                                               \\
\textbf{Scanlan:}                                    & 
\begin{minipage}{0.7\linewidth}\begin{tabularx}{\linewidth}{X}
 No Larkin. \textcolor{darkgreen}{\em I attempt to see see if Tooma is coming.} I don't want to release this thing before Tooma is there reporting to Vorugal.\end{tabularx}\end{minipage}  \\
\textbf{Vaxildan:}                                    & 
\begin{minipage}{0.7\linewidth}\begin{tabularx}{\linewidth}{X}
 (Grog voice) Six. It said six.\end{tabularx}\end{minipage} 
\\ \hline

\end{tabular}

\caption{A sample from CRD3 extended, showing: pairwise character relationships; historical context via the summary; and current character interactions in the form of dialogue, \textcolor{darkgreen}{\emph{first-person} (green)}, and \textcolor{blue}{\underline{second-person} (blue)} narration. DM refers to the Dungeon Master who provides arbitration and additional context to players.}
\label{tab:crd3example1}
\end{table}

We are inspired by the idea of {\em character-driven and emergent storytelling} wherein narrative emerges through characters' interactions as seen in Table~\ref{tab:crd3example1}.
In addition to the challenges faced by automated storytellers, a character-driven storytelling system must produce language while simultaneously: (1) keeping each character's personas consistent while acting; (2) keeping track of relationships between characters that will affect their interactions; and (3) follow and logically advance the plot of the story.

To better explore how to give automated systems these two abilities, we focus on the task of {\em story continuation} solely through dialogue---i.e. picking the next character response that best continues a story. 
The task and data are seen in Table~\ref{tab:crd3example1}.
We build off the Critical Role Dungeons and Dragons Dataset or CRD3~\citep{rameshkumar-bailey-2020-storytelling}, a unique dataset that contains dialogue transcripts of a small group of around six players role-playing various characters while playing the table top role-playing game Dungeons and Dragons---their adventures and interactions forming a narrative that stretches hundreds of chapters, with each chapter forming a subplot.
The original dataset was intended to be used for abstractive summarization and contains ground-truth summaries for each chapter.
To better suit our purpose of studying {\em character-driven storytelling}, we automatically augment the dataset with information regarding character persona as well as relationship types between pairs of characters (friends, enemies, etc.) by clustering crowdsourced descriptions of character interactions from the Critical Role Wiki.\footnote{\url{https://criticalrole.fandom.com/wiki/Critical_Role_Wiki}}

This extended dataset lets us break down the problem of {\em character-driven story continuation} into two sub-tasks corresponding to the three challenges mentioned earlier in terms of interacting within the confines of a story while staying consistent with respect to character personas and relationships.
We show that training a system to optimize for both of these sub-tasks significantly improves story continuation accuracy.

Our work's two primary contributions are thus: (1) the extension to CRD3 enabling a study of {\em character-driven storytelling} and the corresponding methodology used; and (2) a multi-task learning system that leverages character relation and persona information to better complete stories.

\section{Related Work and Background}
\paragraph{Storytelling.}
Storytelling systems that use symbolic planning~\citep{Lebowitz1987,Gervas2005,Porteous2009,Riedl2010a,ware11} focused on ensuring coherence and consistency of plot through explicitly listed rules in the form of pre- and post-conditions, often requiring extensive knowledge engineering.
Modern neural language-model based approaches generally attempt to learn to tell {\em plot-driven stories} from a corpus of stories via learning objectives that optimize reconstructing the story itself~\citep{kiros2015skip, Roemmele2018b,Khalifa2017,Fan2018}.
In particular, a two-step process in which the high level plot is first generated, followed by filling in rest of the story constrained to the plot has emerged~\citep{Martin2017a, martin2018, ammanabrolu2019story,Tambwekar2019,Yao, ippolito-etal-2019-unsupervised}.
\citet{ammanabrolu2020automated} look at plot generation from a {\em character-driven} perspective using commonsense knowledge, though do not model character interactions at all.
Closely related to the spirit of our task is the Story Cloze test~\citep{Mostafazadeh2016c}, which measures the ability of a model to correctly predict the end of a story.
Like the other works mentioned here, however, this task does not require dialogue or other forms of character interactions.

\paragraph{Dialogue.}
Contemporary neural dialogue retrieval systems, both chit-chat and goal-oriented, more explicitly model agent interactions than most storytelling systems~\citep{henderson2014second,asri2017frames}.
Particularly relevant to our work are dialogue systems that attempt to model and stay consistent with an agent's persona, such as Persona Chat~\citep{zhang-etal-2018-personalizing}, or using further contextual information such as setting in addition to character personas using a crowd-sourced fantasy text-game such as LIGHT~\citep{urbanek2019light}.
None of these works, however, have any notion of story or plot, often using significantly less long-term context than most storytelling systems.

\section{Character-Driven Storytelling}
This section first describes the automated extensions to the CRD3 dataset, specifically information on character relationships, followed by the multi-task learning setup and transformer architecture that leverage the new data for story continuation.
\begin{table}[h]
\footnotesize

\centering
\begin{tabular}{|l|r|r|r|}
\hline
                          & \textbf{Train} & \textbf{Valid} & \textbf{Test}  \\ \hline
Avg. no. of turns in a chunk & 38.37 & 61.17 & 62.18 \\ \hline
Avg. no. of char.s in a chunk & 4.06  & 4.07  & 4.36  \\ \hline
No. of chunks              & 11400 & 815   & 761   \\ \hline
\end{tabular}
\caption{CRD3 (extended) dataset statistics.}
\label{tab:crd3stats}
\end{table}

\subsection{CRD3 Automated Dataset Extension}
\label{sec:dataext}
CRD3, as originally seen in \citet{rameshkumar-bailey-2020-storytelling}, contains two seasons of 159 transcribed Critical Role episodes, consisting of 398,682 turns in total.
It further contains 34,243 ground truth human-written summary dialogue chunks that abstractively summarize dialogue chunks.
The chunks themselves consist of a sequence of dialogue and first- and second-person narration turns that form a semantically cohesive unit---with the end of a chunk signifying the completion of a sub-plot or change in location.
Table~\ref{tab:crd3stats} provides statistics for the number of chunks in the train, as well as the average number of character turns and number of characters within a chunk.

To enable a more effective study of {\em character-driven storytelling} using this dataset, we automatically extend CRD3 by adding descriptions of character relations from the Critical Role Wiki.
These descriptions are free form text and often summarize character emotions during their interactions with another character.
To condense them down, we cluster the character relation descriptions in an unsupervised fashion by calculating the vectorized TF-IDF representation of the description and applying the K-means algorithm. 
Varying the number of clusters changes the qualitative information conveyed by the cluster.
For example, if we set the number of clusters to three, we can then also use the popular sentiment analysis tool VADER~\citep{Hutto2014VADERAP} to provide human interperable relationship labels for each of the three clusters---positive, negative, or neutral as seen in Table~\ref{tab:crd3example1}.
We specifically focus on incorporating these 3 relation types into our models.
These relationship labels are attached to every dialogue chunk based on the characters appearing in that chunk.
Further information regarding clusters is found in Appendix~\ref{app:clustering}.

\begin{figure}
\centering
\includegraphics[width=0.5\textwidth]{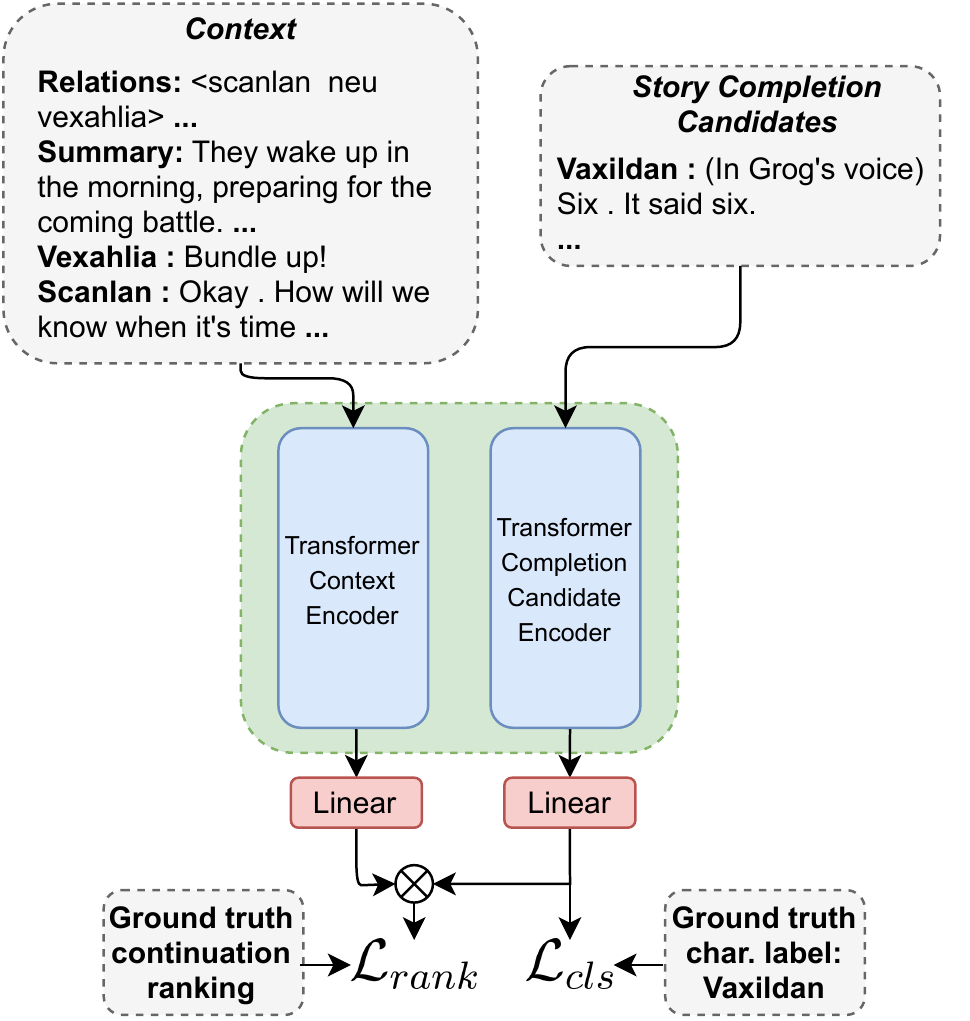}
\caption{Multi-task learning overall architecture. The red shaded linear layers are task-specific and blue transformer blocks are pre-trained. Both transformer blocks share parameters across tasks.}
\label{fig:arch}
\end{figure}

\begin{table*}[t]
\centering
\footnotesize
\begin{tabular}{|l|l|l||l|l|l|l|}
\hline
{\textbf{Eval Task}} & \multicolumn{2}{c||}{\textbf{Character Prediction}} & \multicolumn{4}{c|}{\textbf{Story Continuation}} \\ \hline
{\textbf{Metric}} & \multicolumn{2}{c||}{\textbf{Weighted F1}} & \multicolumn{2}{c|}{\textbf{Hits@1/10}} & \multicolumn{2}{c|}{\textbf{Hits@5/10}} \\ \hline
{\textbf{Training Task Type}} & \multicolumn{1}{c|}{\textbf{Single}} & \multicolumn{1}{c||}{\textbf{Multi}} & \multicolumn{1}{c|}{\textbf{Single}} & \multicolumn{1}{c|}{\textbf{Multi}} & \multicolumn{1}{c|}{\textbf{Single}} & \multicolumn{1}{c|}{\textbf{Multi}} \\ \hline
\textbf{Base} & 47.3 & 47.6 & 18.0 & 18.3 & 70.6 & 73.9 \\
\textbf{Base+Summary} & 48.4 & \textbf{49.0} & 18.0 & 20.4 & 71.7 & 74.3 \\
\textbf{Base+Relations} & \textbf{49.0} & 48.8 & 17.6 & 20.2 & 70.6 & 74.0 \\
\textbf{Base+Summary+Relations} & 48.8 & 48.8 & 18.0 & \textbf{21.3} & \textbf{72.9} & \textbf{74.6}\\ \hline
\end{tabular}
\caption{Multi-task ablations.}
\label{tab:multitask}
\end{table*}

\subsection{Multi-task Learning}
\label{sec:multitask}
Based on the hypothesis that modeling character interaction information is critical for our overall task of {\em character-driven storytelling}, our system optimizes for two sub-tasks: next character prediction and story continuation. 
The next character prediction task can be summarized as: given current context, predict the next character who will act or speak---providing a proxy for judging who is most likely to respond to the current character in a multi-character setting.
Similarly, the story continuation task refers to predicting the next character response that continues the story given the same context.
The context itself contains information regarding: 
(1) a summary of the story so far using the dialogue summary chunks provided in CRD3 and described in Section~\ref{sec:dataext}, (2) pairwise relationship cluster labels between all characters within the dialogue chunk, and (3) the last $n$-turns of character interactions.

Our model's architecture is shown in Figure~\ref{fig:arch}.
It is inspired by the bi-encoder featured in \citet{urbanek2019light}.
In this model, two separate transformers are used to produce vector representations for the input context and each candidate utterance for the response retrieval task.
All candidates are scored by via dot product between their vector representations and the context representation and trained using a ranking loss $\mathcal{L}_{rank}$.
For the task of next character prediction, we use the same vector representation for the context and pass it through an additional linear layer with softmax layer to predict the correct character from the list of all possible characters.
This sub-task uses a cross entropy loss $\mathcal{L}_{cls}$.
The entire system is trained jointly by optimizing 
$\mathcal{L} = \lambda_1 \mathcal{L}_{rank} +\lambda_2 \mathcal{L}_{cls}$ for some hyperparameters $\lambda_i$.
By virtue of the architecture, network parameters are shared between the tasks.

\begin{table}[h]
\footnotesize
\centering
\begin{tabular}{|l|l|l|}
\hline
{\textbf{Eval Task}}      & \multicolumn{1}{|c|}{\textbf{Char. Pred.}}  & \multicolumn{1}{|c|}{\textbf{Story Continuation}} \\ \hline
    {\textbf{Metric}}      & \multicolumn{1}{|c|}{\textbf{Weighted F1}}  & \multicolumn{1}{|c|}{\textbf{Hits@1/10}} \\ \hline
1  & 24.2 & 17.0    \\ \hline
2  & 42.6 & 18.8    \\ \hline
5  & 47.2 & 18.2    \\ \hline
10 & 47.6 & 20.5    \\ \hline
\end{tabular}
\caption{Historical context ablations.}
\label{tab:histlen}
\end{table}
\section{Evaluation}
We conduct two ablations studies that analyze: (1) the complexity of performing {\em character-driven story continuation} on the dataset; and (2) the effectiveness of imbuing the model with relation information via input context and multi-task learning. 

Our base transformer model that we build off of in each of these is the bi-encoder ranker described in Section~\ref{sec:multitask}.
The transformer encoder is a similar architecture as BERT~\citep{devlin-etal-2019-bert}, with 256 million total parameters, and is pre-trained using the Reddit dataset extracted and made available on pushshift.io~\citep{baumgartner2020pushshift} seen in \citet{roller2020recipes}.
This dataset has been shown to result in an improved understanding of conversational natural language~\citep{yang2018learning,mazare-etal-2018-training}.
Further hyperparameter and training details are shown in Appendix~\ref{app:hyperparameter}.

For story continuation report standard retrieval metrics of Hits@$N$, where we measure the ability of the model to output the gold standard dialogue candidate in the top-$N$ of the given candidates.
For character prediction, we report F1 weighted by the number of instances of each character type.

\subsection{Historical Context Ablations}

The first set of ablations measures performance on each of the two sub-tasks as a function of historical context required in an attempt to assess the complexity of the CRD3 extended dataset and its suitability for exploring {\em character-driven storytelling}.
Recall that the CRD3 dataset provides summaries for each separate dialogue chunk.
In Table~\ref{tab:histlen}, we vary the number of prior chunks of such summaries used as input context to the model and measure performance on each of the sub-tasks after training the model jointly on both sub-tasks.

The trends shown in Table~\ref{tab:histlen} are quite clear---indicating that, overall, the CRD3 dataset requires very long contexts to ensure effective performance.
On average, across both evaluation tasks performance gain between using a single historical context chunk and using two is greater than the corresponding differences when using even more chunks.
Additionally, performance continues to rise with added historical context up to the maximum context length we tested of 10.
We note that this is a significantly greater amount of context than generally required for state-of-the-art chit-chat dialogue datasets~\citep{roller2020recipes} as well as prior story completion datasets such as ROC Stories~\citep{Mostafazadeh2016c}, reinforcing our hypothesis that the CRD3 dataset is well suited to enabling {\em character-driven storytelling} by focusing on interactions requiring long-term memory.

\subsection{Multi-task Ablations}
These ablations focus on analyzing the effects of our methods to imbue the agent with relationship and character information, specifically including the relationship cluster labels in the input and multi-task training.
Table~\ref{tab:multitask} outlines these results when evaluated on both the character prediction and story continuation sub-tasks with different: (1) inputs types---with base referring to only character interactions and additional information as seen in Figure~\ref{fig:arch}; and (2) training methods---single referring to training on only the evaluation task and multi to jointly training on both tasks.

We would first like to note that we use the same relationship labels for characters through the entire story---i.e. across all the dialogue chunks.
Our approach intuitively averages the relationship type between characters through time---e.g. characters that are friends at first and then become enemies will have a neutral label throughout all the story.
While more fine grained relationship labels that do not perform such averaging might perform better, they would also require extensive additional human annotations to track relationships through time.

For character prediction, the Base+Summary multi-task and Base+Relations single-task models perform best though are closely comparable to the Base+Summary+Relations multi-task model.
For story continuation, the Base+Summary+Relations multi-task model outperforms all others.
In all story continuation experiments, multi-task trained models outperform their counterpart single-task trained model.
Through these results, we can infer that imbuing character relationship information through {\em both} input relationship cluster information as well as next character prediction helps models continue stories---while staying consistent with a particular character's persona---more accurately.

\section{Conclusions}
We hypothesized that injecting models with information on relationships between characters would improve their ability to complete {\em character-driven stories}.
A series of ablation studies support this, with a key insight being that a particularly efficient way of giving story continuation models this information is by multi-task training them on both character dialogue and relationship information automatically extracted from online sources.

\section{Broader Impacts}
Our work on {\em character-driven storytelling} has potential implications extending to the creation of learning agents that communicate using natural language, especially those requiring an agent to stay consistent with a character or persona throughout an interaction.
As our system is trained entirely using a dataset collected from character interactions of a set of players role-playing in a fantasy Dungeons and Dragons world, we are prone to echoing biases found in the data.
Some of these biases are necessary for effective story continuation, enabling a reader to identify the genre and conveying thematic information.
Others may potentially involve non-normative language usage---acceptable in a fantasy world but inappropriate in the real world.
Restricting our system to {\em story continuation} through a retrieval mechanism as opposed to generating text mitigates, though do not eliminate some of these biases.
We urge future researchers and application developers that use automated storytelling techniques to similarly clarify the origins and methodology behind the creation of delivered story content.

\bibliographystyle{acl_natbib}
\bibliography{acl2021.bib,anthology.bib}

\begin{thebibliography}{28}
\expandafter\ifx\csname natexlab\endcsname\relax\def\natexlab#1{#1}\fi

\bibitem[{Ammanabrolu et~al.(2021)Ammanabrolu, Cheung, Broniec, and
  Riedl}]{ammanabrolu2020automated}
Prithviraj Ammanabrolu, Wesley Cheung, William Broniec, and Mark~O Riedl. 2021.
\newblock Automated storytelling via causal, commonsense plot ordering.
\newblock In \emph{Proceedings of the Thirty-Fifth AAAI Conference on
  Artificial Intelligence}.

\bibitem[{Ammanabrolu et~al.(2020)Ammanabrolu, Tien, Cheung, Luo, Ma, Martin,
  and Riedl}]{ammanabrolu2019story}
Prithviraj Ammanabrolu, Ethan Tien, Wesley Cheung, Zhaochen Luo, William Ma,
  Lara~J Martin, and Mark~O Riedl. 2020.
\newblock Story realization: Expanding plot events into sentences.
\newblock In \emph{Proceedings of the Thirty-Fourth AAAI Conference on
  Artificial Intelligence}.

\bibitem[{Baumgartner et~al.(2020)Baumgartner, Zannettou, Keegan, Squire, and
  Blackburn}]{baumgartner2020pushshift}
Jason Baumgartner, Savvas Zannettou, Brian Keegan, Megan Squire, and Jeremy
  Blackburn. 2020.
\newblock The pushshift reddit dataset.
\newblock In \emph{Proceedings of the International AAAI Conference on Web and
  Social Media}, volume~14, pages 830--839.

\bibitem[{Devlin et~al.(2019)Devlin, Chang, Lee, and
  Toutanova}]{devlin-etal-2019-bert}
Jacob Devlin, Ming-Wei Chang, Kenton Lee, and Kristina Toutanova. 2019.
\newblock \href {https://doi.org/10.18653/v1/N19-1423} {{BERT}: Pre-training of
  deep bidirectional transformers for language understanding}.
\newblock In \emph{Proceedings of the 2019 Conference of the North {A}merican
  Chapter of the Association for Computational Linguistics: Human Language
  Technologies, Volume 1 (Long and Short Papers)}, pages 4171--4186,
  Minneapolis, Minnesota. Association for Computational Linguistics.

\bibitem[{El~Asri et~al.(2017)El~Asri, Schulz, Sharma, Zumer, Harris, Fine,
  Mehrotra, and Suleman}]{asri2017frames}
Layla El~Asri, Hannes Schulz, Shikhar Sharma, Jeremie Zumer, Justin Harris,
  Emery Fine, Rahul Mehrotra, and Kaheer Suleman. 2017.
\newblock Frames: a corpus for adding memory to goal-oriented dialogue systems.
\newblock In \emph{Proceedings of the 18th Annual SIGdial Meeting on Discourse
  and Dialogue}, pages 207--219, Saarbr\"ucken, Germany. Association for
  Computational Linguistics.

\bibitem[{Fan et~al.(2018)Fan, Lewis, and Dauphin}]{Fan2018}
Angela Fan, Mike Lewis, and Yann Dauphin. 2018.
\newblock \href {http://arxiv.org/abs/1805.04833} {{Hierarchical Neural Story
  Generation}}.
\newblock In \emph{Proceedings of the 56th Annual Meeting of the Association
  for Computational Linguistics}, pages 889--898.

\bibitem[{Gerv{\'{a}}s et~al.(2005)Gerv{\'{a}}s, D{\'{i}}az-Agudo, Peinado, and
  Herv{\'{a}}s}]{Gervas2005}
Pablo Gerv{\'{a}}s, Bel{\'{e}}n D{\'{i}}az-Agudo, Federico Peinado, and Raquel
  Herv{\'{a}}s. 2005.
\newblock {Story plot generation based on CBR}.
\newblock \emph{Knowledge-Based Systems}, 18(4-5):235--242.

\bibitem[{Henderson et~al.(2014)Henderson, Thomson, and
  Williams}]{henderson2014second}
Matthew Henderson, Blaise Thomson, and Jason~D Williams. 2014.
\newblock The second dialog state tracking challenge.
\newblock In \emph{Proceedings of the 15th Annual Meeting of the Special
  Interest Group on Discourse and Dialogue (SIGDIAL)}, pages 263--272.

\bibitem[{Hutto and Gilbert(2014)}]{Hutto2014VADERAP}
C.~Hutto and E.~Gilbert. 2014.
\newblock Vader: A parsimonious rule-based model for sentiment analysis of
  social media text.
\newblock In \emph{ICWSM}.

\bibitem[{Ippolito et~al.(2019)Ippolito, Grangier, Callison-Burch, and
  Eck}]{ippolito-etal-2019-unsupervised}
Daphne Ippolito, David Grangier, Chris Callison-Burch, and Douglas Eck. 2019.
\newblock \href {https://doi.org/10.18653/v1/W19-2405} {Unsupervised
  hierarchical story infilling}.
\newblock In \emph{Proceedings of the First Workshop on Narrative
  Understanding}, pages 37--43, Minneapolis, Minnesota. Association for
  Computational Linguistics.

\bibitem[{Khalifa et~al.(2017)Khalifa, Barros, and Togelius}]{Khalifa2017}
Ahmed Khalifa, Gabriella A.~B. Barros, and Julian Togelius. 2017.
\newblock \href {http://arxiv.org/abs/1705.03557} {{DeepTingle}}.
\newblock In \emph{International Conference on Computational Creativity}.

\bibitem[{Kiros et~al.(2015)Kiros, Zhu, Salakhutdinov, Zemel, Urtasun,
  Torralba, and Fidler}]{kiros2015skip}
Ryan Kiros, Yukun Zhu, Ruslan~R Salakhutdinov, Richard Zemel, Raquel Urtasun,
  Antonio Torralba, and Sanja Fidler. 2015.
\newblock Skip-thought vectors.
\newblock In \emph{Advances in neural information processing systems}, pages
  3294--3302.

\bibitem[{Lebowitz(1987)}]{Lebowitz1987}
Michael Lebowitz. 1987.
\newblock {Planning Stories}.
\newblock In \emph{Proceedings of the 9th Annual Conference of the Cogntive
  Science Society}, pages 234--242.

\bibitem[{Martin et~al.(2018)Martin, Ammanabrolu, Wang, Hancock, Singh,
  Harrison, and Riedl}]{martin2018}
Lara~J. Martin, Prithviraj Ammanabrolu, Xinyu Wang, William Hancock, Shruti
  Singh, Brent Harrison, and Mark~O. Riedl. 2018.
\newblock {Event Representations for Automated Story Generation with Deep
  Neural Nets}.
\newblock In \emph{Thirty-Second AAAI Conference on Artificial Intelligence
  (AAAI-18)}, pages 868--875, New Orleans, Louisiana.

\bibitem[{Martin et~al.(2017)Martin, Ammanabrolu, Wang, Singh, Harrison,
  Dhuliawala, Tambwekar, Mehta, Arora, Dass, Purdy, and Riedl}]{Martin2017a}
Lara~J. Martin, Prithviraj Ammanabrolu, Xinyu Wang, Shruti Singh, Brent
  Harrison, Murtaza Dhuliawala, Pradyumna Tambwekar, Animesh Mehta, Richa
  Arora, Nathan Dass, Chris Purdy, and Mark~O. Riedl. 2017.
\newblock \href
  {https://nips2017creativity.github.io/doc/Improvisational{\_}Agents.pdf}
  {{Improvisational Storytelling Agents}}.
\newblock In \emph{Workshop on Machine Learning for Creativity and Design
  (NeurIPS 2017)}, Long Beach, CA.

\bibitem[{Mazar{\'e} et~al.(2018)Mazar{\'e}, Humeau, Raison, and
  Bordes}]{mazare-etal-2018-training}
Pierre-Emmanuel Mazar{\'e}, Samuel Humeau, Martin Raison, and Antoine Bordes.
  2018.
\newblock \href {https://doi.org/10.18653/v1/D18-1298} {Training millions of
  personalized dialogue agents}.
\newblock In \emph{Proceedings of the 2018 Conference on Empirical Methods in
  Natural Language Processing}, pages 2775--2779, Brussels, Belgium.
  Association for Computational Linguistics.

\bibitem[{Mostafazadeh et~al.(2016)Mostafazadeh, Chambers, He, Parikh, Batra,
  Vanderwende, Kohli, and Allen}]{Mostafazadeh2016c}
Nasrin Mostafazadeh, Nathanael Chambers, Xiaodong He, Devi Parikh, Dhruv Batra,
  Lucy Vanderwende, Pushmeet Kohli, and James Allen. 2016.
\newblock \href {http://arxiv.org/abs/1604.01696} {{A Corpus and Evaluation
  Framework for Deeper Understanding of Commonsense Stories}}.
\newblock In \emph{Proceedings of the 2016 Conference of the North American
  Chapter of the Association for Computational Linguistics: Human Language
  Technologies}, pages 839--849.

\bibitem[{Porteous and Cavazza(2009)}]{Porteous2009}
Julie Porteous and Marc Cavazza. 2009.
\newblock \href {http://arxiv.org/abs/9780201398298} {{Controlling narrative
  generation with planning trajectories: The role of constraints}}.
\newblock In \emph{Joint International Conference on Interactive Digital
  Storytelling}, volume 5915 LNCS, pages 234--245. Springer.

\bibitem[{Rameshkumar and Bailey(2020)}]{rameshkumar-bailey-2020-storytelling}
Revanth Rameshkumar and Peter Bailey. 2020.
\newblock \href {https://doi.org/10.18653/v1/2020.acl-main.459} {Storytelling
  with dialogue: {A} {Critical Role Dungeons and Dragons Dataset}}.
\newblock In \emph{Proceedings of the 58th Annual Meeting of the Association
  for Computational Linguistics}, pages 5121--5134, Online. Association for
  Computational Linguistics.

\bibitem[{Riedl and Young(2010)}]{Riedl2010a}
Mark~O Riedl and R~Michael Young. 2010.
\newblock \href {https://www.cc.gatech.edu/{~}riedl/pubs/jair.pdf} {{Narrative
  Planning: Balancing Plot and Character}}.
\newblock \emph{Journal of Artificial Intelligence Research}, 39:217--267.

\bibitem[{Roemmele and Gordon(2018)}]{Roemmele2018b}
Melissa Roemmele and Andrew~S Gordon. 2018.
\newblock \href {http://aclweb.org/anthology/W18-1506
  http://people.ict.usc.edu/{~}gordon/publications/NAACL-WS18A.PDF} {{An
  Encoder-decoder Approach to Predicting Causal Relations in Stories}}.
\newblock In \emph{Proceedings of the First Workshop on Storytelling}, pages
  50--59, New Orleans, Louisiana. Association for Computational Linguistics.

\bibitem[{Roller et~al.(2020)Roller, Dinan, Goyal, Ju, Williamson, Liu, Xu,
  Ott, Shuster, Smith et~al.}]{roller2020recipes}
Stephen Roller, Emily Dinan, Naman Goyal, Da~Ju, Mary Williamson, Yinhan Liu,
  Jing Xu, Myle Ott, Kurt Shuster, Eric~M Smith, et~al. 2020.
\newblock Recipes for building an open-domain chatbot.
\newblock \emph{arXiv preprint arXiv:2004.13637}.

\bibitem[{Tambwekar et~al.(2019)Tambwekar, Dhuliawala, Martin, Mehta, Harrison,
  and Riedl}]{Tambwekar2019}
Pradyumna Tambwekar, Murtaza Dhuliawala, Lara~J. Martin, Animesh Mehta, Brent
  Harrison, and Mark~O. Riedl. 2019.
\newblock \href {https://www.ijcai.org/proceedings/2019/829} {{Controllable
  Neural Story Plot Generation via Reward Shaping}}.
\newblock In \emph{Proceedings of the 28th International Joint Conference on
  Artificial Intelligence}.

\bibitem[{Urbanek et~al.(2019)Urbanek, Fan, Karamcheti, Jain, Humeau, Dinan,
  Rocktäschel, Kiela, Szlam, and Weston}]{urbanek2019light}
Jack Urbanek, Angela Fan, Siddharth Karamcheti, Saachi Jain, Samuel Humeau,
  Emily Dinan, Tim Rocktäschel, Douwe Kiela, Arthur Szlam, and Jason Weston.
  2019.
\newblock Learning to speak and act in a fantasy text adventure game.
\newblock In \emph{EMNLP}.

\bibitem[{Ware and Young(2011)}]{ware11}
Stephen Ware and R.~Michael Young. 2011.
\newblock Cpocl: A narrative planner supporting conflict.
\newblock In \emph{Proceedings of the 7th AAAI Conference on Artificial
  Intelligence and Interactive Digital Entertainment}.

\bibitem[{Yang et~al.(2018)Yang, Yuan, Cer, Kong, Constant, Pilar, Ge, Sung,
  Strope, and Kurzweil}]{yang2018learning}
Yinfei Yang, Steve Yuan, Daniel Cer, Sheng-Yi Kong, Noah Constant, Petr Pilar,
  Heming Ge, Yun-Hsuan Sung, Brian Strope, and Ray Kurzweil. 2018.
\newblock Learning semantic textual similarity from conversations.
\newblock \emph{arXiv preprint arXiv:1804.07754}.

\bibitem[{Yao et~al.(2019)Yao, Peng, Weischedel, Knight, Zhao, and Yan}]{Yao}
Lili Yao, Nanyun Peng, Ralph Weischedel, Kevin Knight, Dongyan Zhao, and Rui
  Yan. 2019.
\newblock \href {http://arxiv.org/abs/1811.05701v3} {{Plan-And-Write: Towards
  Better Automatic Storytelling}}.
\newblock In \emph{Proceedings of the Thirty-Third AAAI Conference on
  Artificial Intelligence (AAAI-19)}.

\bibitem[{Zhang et~al.(2018)Zhang, Dinan, Urbanek, Szlam, Kiela, and
  Weston}]{zhang-etal-2018-personalizing}
Saizheng Zhang, Emily Dinan, Jack Urbanek, Arthur Szlam, Douwe Kiela, and Jason
  Weston. 2018.
\newblock \href {https://doi.org/10.18653/v1/P18-1205} {Personalizing dialogue
  agents: {I} have a dog, do you have pets too?}
\newblock In \emph{Proceedings of the 56th Annual Meeting of the Association
  for Computational Linguistics (Volume 1: Long Papers)}, pages 2204--2213,
  Melbourne, Australia. Association for Computational Linguistics.

\end{thebibliography}

\clearpage

\appendix
\onecolumn
\section{Appendices}
\subsection{CRD3 Extended Examples}
\label{app:crd3}
\begin{table}[h]
\centering
\begin{tabular}{|rl|}
\hline
\textbf{Relations} & \begin{tabular}[c]{@{}l@{}}\textless{}Scanlan, neutral, Vexahlia\textgreater{},\\ \textless{}Grog, neutral, Scanlan\textgreater{}, ...\end{tabular}                                                                                                                                                                                              \\ \hline \hline
\textbf{Summary}   & \begin{tabular}[c]{@{}l@{}}Scanlan deceives the clasp leader with a blue gem that can grant one \\ wish if they say the password while holding the gem. He gives the \\ leader the gem and promises to give him the password if they can \\ visit riskel. The leader reveals the clasp helped riskel prepare \\ for his escape. ...\end{tabular} \\ \hline \hline
\textbf{Keyleth:}  & okay !                                                                                                                                                                                                                                                                                                                                           \\
\textbf{DM:}       & \begin{tabular}[c]{@{}l@{}}He looks over at the gentleman who inspected it earlier and nods his \\ head. "accepted." and they continue walking forward.\end{tabular}                                                                                                                                                                        \\
\textbf{Grog:}     & Lucky fucking druid.                                                                                                                                                                                                                                                                                                                            \\
\textbf{DM:}       & It is the piece you put in the actual--                                                                                                                                                                                                                                                                                                          \\
\textbf{Scanlan:}  & \begin{tabular}[c]{@{}l@{}}It's a blue shard that we found in--long, long ago-- it's real crystal \\ and it's real magic.\end{tabular}                                                                                                                                                                                                     \\
\textbf{DM:}       & Yes. I know what that is.                                                                                                                                                                                                                                                                                                                      \\
\textbf{Scanlan:}  & Because I don't.                                                                                                                                                                                                                                                                                                                               \\
\textbf{DM:}       & Well, it was sufficient upon inspection for this.                                                                                                                                                                                                                                                                                              \\
\textbf{Scanlan:}  & Okay.                                                                                                                                                                                                                                                                                                                                           \\
\textbf{Vexahlia:} & Whoa, I think it opens a portal to another plane.                                                                                                                                                                                                                                                                                   \\
\textbf{Scanlan:}  & I don't know what it is, but it's magic.                                                                                                                                                                                                                                                                                                     \\ \hline
\end{tabular}
\caption{Randomly selected CRD3 extended examples}
\end{table}

\begin{table}[h]
\centering
\begin{tabular}{|rl|}
\hline
\textbf{Relations} & \begin{tabular}[c]{@{}l@{}}\textless{}Grog, neutral, Vexahlia\textgreater{},\\ \textless{}Keyleth, positive, Scanlan\textgreater{}, ...\end{tabular}                                                                                                                                        \\ \hline \hline
\textbf{Summary}   & \begin{tabular}[c]{@{}l@{}}Rejoining the party, Vex wonders aloud why desmond is still in the \\ cell. Percy responds that it was originally for his own protection, \\ but that since the problem has been taken care of, it is a precaution \\ that is no longer needed. ...\end{tabular} \\ \hline \hline
\textbf{Vexahlia:} & Are there days of the week? what is a weekend?                                                                                                                                                                                                                                            \\
\textbf{Keyleth:}  & Yeah, There's days of the week .                                                                                                                                                                                                                                                          \\
\textbf{Scanlan:}  & What is this world? How does time work here?                                                                                                                                                                                                                                              \\
\textbf{DM:}       & \begin{tabular}[c]{@{}l@{}}There are days of the week, I'm not gon na go into the specifics of \\ it because I'm working on it. This question hasn't really arisen \\ before and I probably should figure that out. It 's the equivalent of \\ a thursday.\end{tabular}              \\
\textbf{Scanlan:}  & It's always thursday.                          \\ \hline
\end{tabular}
\caption{Randomly selected CRD3 extended examples}
\end{table}

\onecolumn
\subsection{Experiment parameter details}
\label{app:hyperparameter}

\begin{table}[h]
\centering
\begin{tabular}{|l|l|}
\hline
\multicolumn{2}{|c|}{Model Parameters} \\ \hline
no. clusters used            & 3    \\ \hline
attention dropout            & 0.2    \\ \hline
dropout                       & 0.1    \\ \hline
learning rate                & 5e-05  \\ \hline
learning rate decay rate   & 0.4    \\ \hline
learning rate warmup steps & 100    \\ \hline
number of epochs                    & 10     \\ \hline
optimizer                     & adamax \\ \hline
number of heads                    & 12     \\ \hline
hidden layers           & 12     \\ \hline
hidden size                  & 3072   \\ \hline
embedding size                & 768    \\ \hline
batch size                    & 10     \\ \hline
activation                    & gelu   \\ \hline
$\lambda_1$                     & 0.5   \\ \hline
$\lambda_2$                     & 0.5   \\ \hline
\end{tabular}
\caption{Hyperparameters for training the multi-task model. Parameters were adapted from \citet{urbanek2019light} and not tuned further. All experiments were conducted on 4 Nvidia TitanX GPUs and take less than 24 hours.}
\end{table}

\subsection{Clustering Examples}
\label{app:clustering}
\begin{table}[h]
\scalebox{0.75}{
\begin{tabular}{|llcccc|}
\hline
Character pair      & Description of relationship                                                                                                                                                                                                                                                            & \multicolumn{1}{l}{cluster 2} & \multicolumn{1}{l}{\#cluster 3} & \multicolumn{1}{l}{\#cluster 24} & VADER \\ \hline
(Percy,Scanlan)     & \begin{tabular}[c]{@{}l@{}}Although Scanlan and Percy didn't interact\\ much, for the most part, the two had an amicable \\ relationship.\end{tabular}                                                                                                                                 & 1                             & 0                               & 6                                & neu                         \\
(Grog,Trinket)      & \begin{tabular}[c]{@{}l@{}}At some point, Grog almost got Trinket killed \\ when he hit the bear's backside with the flat of \\ his axe, startling him and sending him charging \\ through a hallway full of traps, triggering every \\ one of them in the process.\end{tabular}       & 1                             & 1                               & 5                                & neg                        \\
(Pike,Vex'ahlia)    & \begin{tabular}[c]{@{}l@{}}Pike and Vex have an incredibly fond and comfortable \\ rapport with each other.\end{tabular}                                                                                                                                                               & 0                             & 1                               & 17                               & pos                       \\
(Percy,Pike)        & \begin{tabular}[c]{@{}l@{}}Pike and Percy are good friends and have the utmost \\ respect for each other, despite Percy's distrust of the gods.\end{tabular}                                                                                                                           & 0                             & 1                               & 23                               & pos                         \\
(Keyleth,Scanlan)   & \begin{tabular}[c]{@{}l@{}}At some point during the adventures of Vox Machina, \\ Keyleth was thrown in jail. Scanlan managed to get her \\ out by convincing the guards that she had pubic lice. \\ The druid played along by foaming at the mouth and \\ acting insane.\end{tabular} & 1                             & 2                               & 10                               & pos                         \\
(Trinket,Vex'ahlia) & \begin{tabular}[c]{@{}l@{}}Vex'ahlia and Trinket are close companions, having \\ traveled together since Trinket was a cub.\end{tabular}                                                                                                                                               & 0                             & 2                               & 4                                & neu                       \\ \hline
\end{tabular}
}
\caption{Pair-wise character relationship information using clustering and VADER.}
\label{app:cluster}
\end{table}

\end{document}